\documentclass[a4paper]{article}
\usepackage[pdftex]{graphicx}
\usepackage{multirow}
\usepackage{amssymb,amsmath,bm}
\newcommand{\argmax}{\mathop{\rm arg~max}\limits}
\newcommand{\argmin}{\mathop{\rm arg~min}\limits}
\usepackage{cite}
\usepackage{cases}
\usepackage{nidanfloat}
\usepackage{multirow}

\newcommand{\vs}{\vspace{-1pt}}
\newcommand{\svs}{\vspace{0pt}}

\newcommand{\vstmp}{\vspace{-0.5pt}}
\usepackage{INTERSPEECH2021}
\title{End-to-End Rich Transcription-Style Automatic Speech Recognition \\with Semi-Supervised Learning}
\name{\begin{tabular}{c}
    Tomohiro Tanaka, Ryo Masumura, Mana Ihori, \\Akihiko Takashima, Shota Orihashi, Naoki Makishima\end{tabular}}
\address{NTT Media Intelligence Laboratories, NTT Corporation}
\email{tomohiro.tanaka.ht@hco.ntt.co.jp}
\begin{document}
\maketitle
\begin{abstract}
We propose a semi-supervised learning method for building end-to-end rich transcription-style automatic speech recognition (RT-ASR) systems from small-scale rich transcription-style and large-scale common transcription-style datasets. In spontaneous speech tasks, various speech phenomena such as fillers, word fragments, laughter and coughs, etc. are often included. While common transcriptions do not give special awareness to these phenomena, rich transcriptions explicitly convert them into special phenomenon tokens as well as textual tokens. In previous studies, the textual and phenomenon tokens were simultaneously estimated in an end-to-end manner. However, it is difficult to build accurate RT-ASR systems because large-scale rich transcription-style datasets are often unavailable. To solve this problem, our training method uses a limited rich transcription-style dataset and common transcription-style dataset simultaneously. The Key process in our semi-supervised learning is to convert the common transcription-style dataset into a pseudo-rich transcription-style dataset. To this end, we introduce style tokens which control phenomenon tokens are generated or not into transformer-based autoregressive modeling. We use this modeling for generating the pseudo-rich transcription-style datasets and for building RT-ASR system from the pseudo and original datasets. Our experiments on spontaneous ASR tasks showed the effectiveness of the proposed method.
\end{abstract}
\noindent\textbf{Index Terms}: automatic speech recognition, rich transcription, semi-supervised learning, pseudo-labeling
\section{Introduction}
\label{sec:intro}
There has been increasing progress in end-to-end automatic speech recognition (ASR) that directly converts a speech into textual tokens (characters, words, etc.).
End-to-end ASR is an approach with simpler training and faster decoding compared with traditional deep-neural-network hidden-Markov-model hybrid ASR.
Various end-to-end ASR methods including connectionist temporal classification \cite{DBLP:conf/icml/GravesJ14,DBLP:journals/corr/HannunCCCDEPSSCN14,DBLP:conf/asru/MiaoGM15}, recurrent neural network (RNN) encoder-decoder models \cite{Chorowski_nips15,is15_rnn_asr,DBLP:conf/icassp/BahdanauCSBB16,listen_and}, RNN transducers \cite{graves_transd13,transd_asru17}, and transformer encoder-decoder models \cite{sptransf,all_you_need}, have been investigated.
These methods have been successful in recognizing textual information in speech.

General ASR systems usually transcribe speech into only textual information, but spontaneous speech includes rich information.
For example, spontaneous speech often includes speech phenomena such as fillers, word fragments, laughter, and coughs.
Since the speech changes due to the occurrence of such phenomena, ASR errors are induced in general ASR systems.
Laughter obscures speech, and fillers and word fragments make connections of textual tokens irregular.
Therefore, it is important to consider these phenomena when estimating textual information for accurate ASR.

In previous studies, speech recognition and detecting speech phenomena were jointly modeled for forming an end-to-end rich transcription-style ASR (RT-ASR) system \cite{inaguma_joint,fujimura_joint}.
Simultaneous estimation of textual information and speech phenomenon was achieved by treating the speech phenomena as phenomenon tokens in the same way as textual tokens.
RT-ASR systems execute speech recognition by taking into account what kind of speech phenomena are occurring and can output textual and phenomenon tokens at the same time.
The models in previous studies were trained from only rich transcription-style datasets including speech, textual tokens, and phenomenon tokens.
However, it is difficult to build highly accurate RT-ASR systems only from the rich transcription-style dataset because collecting large-scale rich transcription-style datasets is costly and time-consuming.

% this study:
In this paper, we propose a semi-supervised learning method for constructing RT-ASR systems using small-scale rich and large-scale common transcription-style datasets consisting of only speech and textual tokens.
In contrast to previous studies, the key advance of our method is to handle both rich and common transcription-style datasets.
In our semi-supervised learning method, we convert the common transcription-style dataset into a pseudo-rich transcription-style dataset and use it for further model training.
Our key idea for handling two different style datasets is to introduce style tokens that indicated rich or common transcription style into transformer-based autoregressive modeling.
By explicitly considering a transcription-style from the style token, the model can switch whether to output the phenomenon tokens.
We apply this modeling to generating a pseudo-rich transcription-style dataset and building a final RT-ASR system.
This RT-ASR system transcribes input speech into a token sequence that includes both textual tokens and phenomenon tokens in an end-to-end manner, the same as previous RT-ASR systems.

There have been several attempts to build an ASR system based on a semi-supervised learning approach \cite{02_ssl_ac,karel_semi,icassp09_semi_lm,icassp19_cycle,icassp20_semi_distil,is20_semi,is20_semi_da}.
For end-to-end ASR, pseudo-labeling methods that generate pseudo-transcriptions against unlabeled data have been increasingly popular \cite{icassp20_semi_distil,is20_semi,is20_semi_da}.
Previous pseudo-labeling methods for end-to-end ASR involve generating pseudo-labels of textual tokens for unlabeled speech.
In our method, we generate pseudo-transcriptions with phenomenon tokens for speech and textual token pairs.
In other words, previous semi-supervised learning methods focused on general end-to-end ASR, whereas our method focuses on end-to-end RT-ASR.

We conduct experiments to evaluate the proposed method on Japanese spontaneous speech recognition tasks.
The experimental results indicate that the proposed method enables better ASR performance in terms of the standard character error rate (CER) and the CER including phenomenon tokens.
Our method enables us to construct highly accurate RT-ASR systems since both a small-scale rich transcription-style and a large-scale common transcription-style datasets can be used for the training by using semi-supervised learning.
We verify that RT-ASR system built with our method outperforms general end-to-end ASR systems when the same amount of data is used.
\section{Rich Transcription-Style Dataset}
\label{sec:data}
\begin{table}[t]
\centering
\caption{Speech phenomena in rich transcription-style dataset. Number of phenomena is that counted in 24-hour dataset.}
\label{tab:tag_detail}
\vs\vs\vs
\begin{tabular}{l|l|r}
\hline
Phenomenon                & Type            & \# phenomena \\ \hline
Fragment                  & single token    & 4,361        \\
Laugh                     & single token    & 4,388        \\
Cough                     & single token    & 73           \\
Sigh                      & single token    & 25           \\
Filler                    & enclosed tokens & 18,558       \\
Repetation                & enclosed tokens & 487          \\
Misstatement              & enclosed tokens & 143          \\
Stretch                   & enclosed tokens & 3,296        \\
Laughing                  & enclosed tokens & 2,334        \\
Hard to hear              & enclosed tokens & 3,176        \\
Emphasis                  & enclosed tokens & 821          \\\hline
\end{tabular}
\vs
\end{table}
A common transcription-style dataset is annotated with textual tokens but not with speech phenomena appearing in spontaneous speech.
To handle various speech phenomena including fillers, repetitions, and laughing utterances, a rich transcription-style dataset in which such phenomena are annotated, is required.
We use our natural two-person dialogue corpus (NTDC) in our experiments.
The two individuals were given the roles of questioner and respondent.
The questioner asked questions on a topic we determined beforehand.
We recorded the dialogues and annotated textual and phenomenon tokens.
The total amount of speech in NTDC is 24 hours, and the total number of textual tokens is 125,376.

It is thought that various types of speech phenomena are included in spontaneous speech.
We roughly divided the phenomena into two types to create a rich transcription-style dataset.
One type can be expressed as a single appearance (single token), and the other is something that can be expressed so as to enclose the textual tokens (enclosed tokens).
Examples include fillers, repetitions, and laughing utterances.
Table \ref{tab:tag_detail} shows the phenomena considered for RT-ASR and their frequency in the data.
We define 4 types of single tokens and 7 types of enclosed tokens.
%----------------------------------------------------
\section{Proposed Semi-Supervised Learning Method for RT-ASR}
\label{sec:semi}
\begin{figure}[t]
  \centering
  \includegraphics[width=0.8\linewidth]{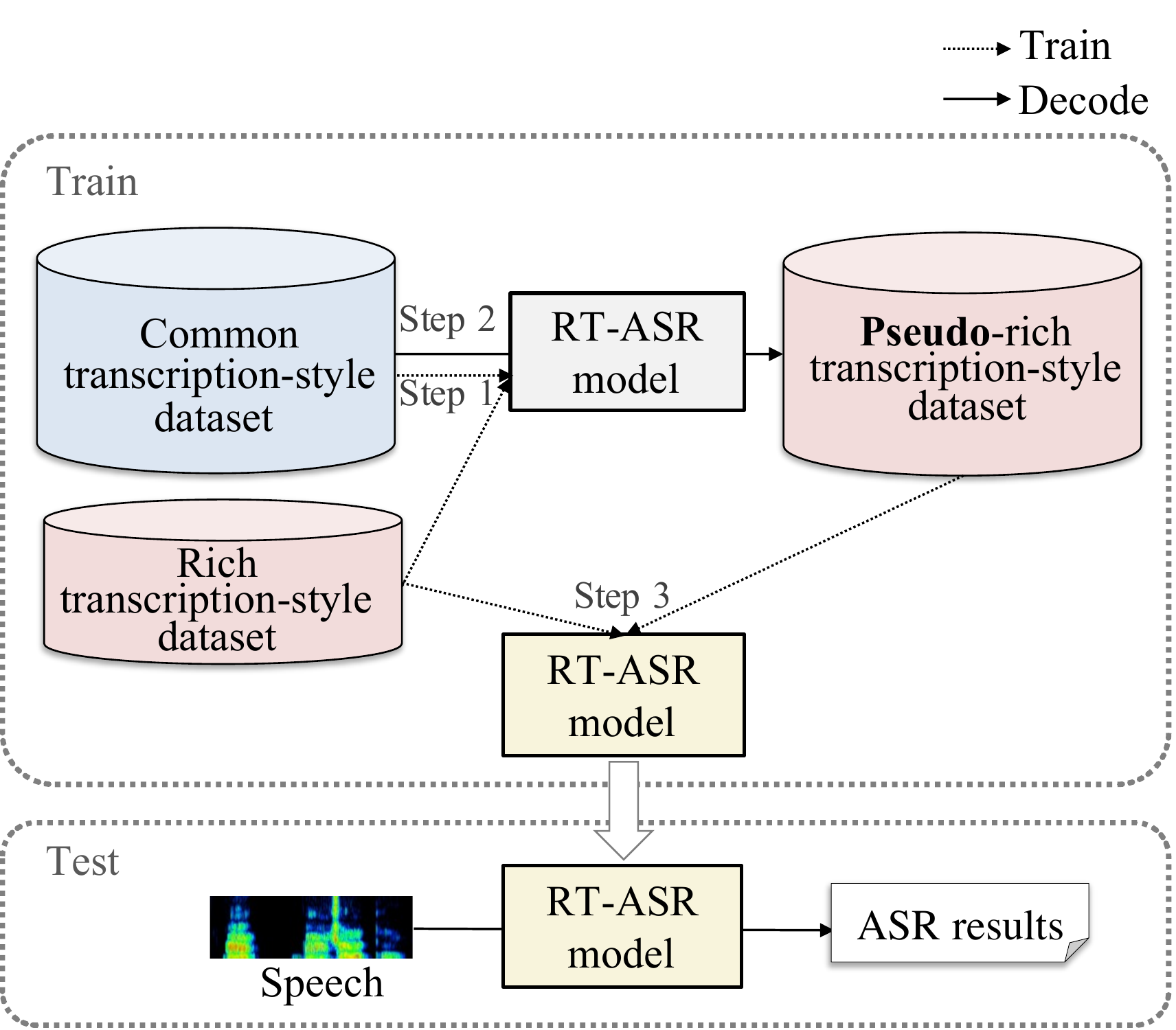}
  \vs
  \caption{Flow of proposed semi-supervised learning methood for RT-ASR.}
  \label{fig:flow}
\end{figure}
\subsection{Strategy}
We take a semi-supervised learning approach to efficiently utilize a small-scale rich transcription-style dataset and large-scale common transcription-style dataset.
Figure \ref{fig:flow} illustrates flow of our semi-supervised learning method for RT-ASR.
In our semi-supervised method, we convert common transcription-style dataset into pseudo-rich transcription-style dataset.
The final model is trained using the pseudo-rich and rich transcription-style datasets.

For the semi-supervised learning, we introduce style tokens that indicate the rich or common transcription-style into transformer-based autoregressive modeling.
The modeling is used for generating pseudo-rich transcription-style dataset and building a final RT-ASR system.
In training, we use a rich transcription-style dataset with a rich-style style token and a common transcription-style dataset with a common-style style token.
The style tokens allow the model to learn whether to output the phenomenon tokens.
In decoding, the model can output rich-style transcription by feeding the rich-style style token.
This enables us to achieve efficient conversion of common transcription-style into rich transcription style.
%----------------------------------------------------
\subsection{Modeling}
\begin{figure}[t]
  \centering
  \includegraphics[width=0.80\linewidth]{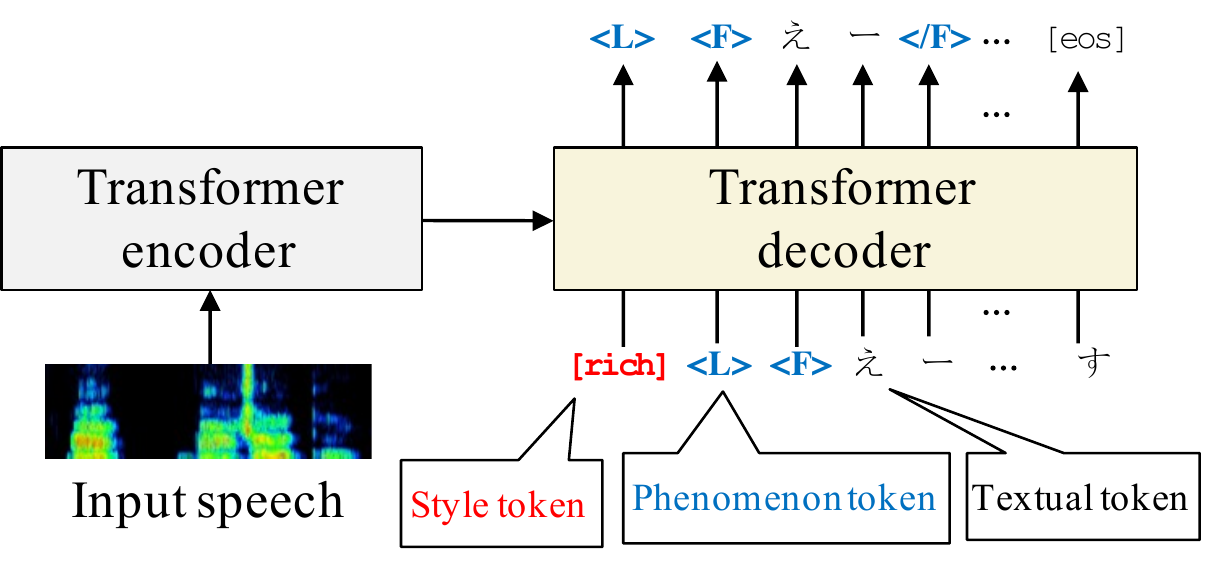}
  \vs
  \caption{End-to-end rich transcrption-style automatic speech recognition. This is the example of inference with style token of rich transcription-style.}
  \label{fig:asr}
  \vs\vs\vs
\end{figure}
Our model directly converts acoustic features of input speech into a sequence of mixed textual tokens and phenomenon tokens.
Given speech $\bm{X} = \{\bm{x}_1, \cdots ,\bm{x}_I\}$ and style token $s \in \{\tt{{[rich]}, \tt{[common]}} \}$ that represent rich transcription-style or common transcription-style, the encoder-decoder estimates the generative probability of a token sequence $\bm{W} = \{w_1, \cdots ,w_T\}$, where $I$ is the number of the acoustic features in the input speech and $T$ is the number of the tokens in the token sequence.
The generative probability of a sequence is defined as
\begin{align}
  P(\bm{W}|\bm{X}, s; \bm{\Theta}) = \prod_{t=1}^T P(w_t| w_{1:t-1}, \bm{X}, s;\bm{\Theta}),
  %s \in {[rich], [common]}
  \label{eq:prob_r}
\end{align}
where $\bm{\Theta}$ represents the trainable parameters. % in the transformer encoder-decoder.

We use transformer-based autoregressive models that predict the probability of a token given the previous predicted tokens.
The encoder converts the input acoustic features $\bm{X}$ into hidden representations $\bm{f}^M$ by using $M$ transformer encoder blocks.
The $m$-th transformer encoder block composes the $m$-th hidden representations $\bm{f}^m$ from the lower layer inputs  $\bm{f}^{m-1}$ as
\begin{align}
  \bm{f}^m & = \mathtt{TransformerEncoderBlock}(\bm{f}^{m-1}; {\theta}_f),
  \label{eq:trenc}
\end{align}
where $\mathtt{TransformerEncoderBlock}(\cdot)$ is a transformer encoder including a scaled dot-product multi-head self-attention layer and a position-wise feed-forward network and ${\theta}_f$ is a trainable parameter. The first input for the encoder $\bm{f}^0 = \{ \bm{\overline{f}}_1, \cdots \bm{\overline{f}}_{I'}\}$ is computed from the input acoustic features as
\begin{align}
  \bm{\overline{f}}_{i'} = \mathtt{PositionalEncoding}(\bm{\overline{x}}_{i'}), \\
  \{ \bm{\overline{x}}_{1}, \cdots , \bm{\overline{x}}_{I'} \}  = \mathtt{Convolution}( \bm{x}_{1}, \cdots ,\bm{x}_{I}; {\theta}_c),
\label{eq:spemb}
\end{align}
where $\mathtt{PositionalEncoding}(\cdot)$ is a function that adds a continuous vector in which position information is embedded, $\mathtt{Convolution}(\cdot)$ is a convolutional neural network consisting of convolutional and pooling layers, and ${\theta}_x$ is the trainable parameter.

The hidden representations of the final block $\bm{f}^M$ in the encoder is fed into the transformer decoder. When the output of the $t$-th time step for the $n$-th transformer block in the decoder is $\bm{g}_{t}^{n}$,
the transformer decoder constructs a hidden representation from the lower output of the decoder $\bm{g}_{0:t-1}^{n-1} = \{ \bm{g}_{0}^{n-1}, \cdots, \bm{g}_{t-1}^{n-1} \}$ as 
\begin{align}
  \bm{g}_{t}^n & = \mathtt{TransformerDecoderBlock}(\bm{g}_{0:t-1}^{n-1}; \bm{f}^M, {\theta}_g),
  \label{eq:trdec}
\end{align}
where $\mathtt{TransformerDecoderBlock}(\cdot)$ is a transformer decoder including a scaled dot-product multi-head masked self-attention layer,
position-wise feed-forward network, and scaled dot product multi-head source-target attention layer, and ${\theta}_g$ is the trainable parameter.
The input of the first transformer block is token embedding, which is calculated as
\begin{align}
   \bm{g}^{0}_{t} & = \mathtt{PositionalEncoding}(\overline{\bm{w}}_{t}),\\
   \overline{\bm{w}}_{t} & = \mathtt{Embedding}(w_t; {\theta}_w),
\end{align}
where $\mathtt{Embedding}(\cdot)$ is a function that converts a token into a continuous representation and ${\theta}_w$ is the trainable parameter.
In a typical encoder-decoder-based ASR model, the first token input into the decoder is $\tt{[sos]}$ that represent the start-of-sentence.
We use special style tokens $s$ as the first input.
In short, the first input is written as $w_0 = s$.
Finally, the network estimates the probabilities of a distribution of the output tokens as
\begin{align}
  P(w_t|w_{1:t-1}; \bm{\Theta}) = \mathtt{Softmax}(\bm{g}_{0:t-1}^{N}; {\theta}_o),
  \label{eq:softmax}
\end{align}
where $\mathtt{Softmax}(\cdot)$ represents the softmax function with linear Transformation, $N$ is the number of blocks in the transformer decoder, and ${\theta}_o$ is the trainable parameter.
The model parameters can be summarized as $\bm{\Theta} = \{{\theta}_f,{\theta}_g,{\theta}_c,{\theta}_w,{\theta}_o\}$.
%---------------------------------------------------
\subsection{Semi-supervised learning}
\noindent {\bf Step 1:} First, we train a model to convert common transcription-style dataset into pseudo-rich transcription-style dataset.
We use common transcription-style dataset $\mathcal{D}_{\rm c}$ and rich transcription-style dataset $\mathcal{D}_{\rm r}$ for the training.
The model parameters $\bm{\Theta}$ are optimized so as to maximize the generative probability in the decoder when given an input speech and a style token.
Thus, the model parameters are optimized by minimizing the cross entropy loss function as
\begin{multline}
  \hat{\bm{\Theta}} = \argmin_{\bm{\Theta}} \\
  - \sum_{\mathcal{D} \in {\{\mathcal{D}_{\rm c},\mathcal{D}_{\rm r}\}}} \sum_{(\bm{W}^{\prime}, \bm{X}^{\prime}) \in {\mathcal{D}}} \log P(\bm{W}^{\prime}|\bm{X}^{\prime}, s; \bm{\Theta}), 
\end{multline}
\begin{subnumcases}
  {s =}
  {\tt [rich]} \text{~ ~ ~ if $\mathcal{D} = \mathcal{D}_{\rm r}$,} & \\
  {\tt [common]} \text{~ ~ ~ if $\mathcal{D} = \mathcal{D}_{\rm c}$,} &
\end{subnumcases}
where $\mathcal{D}$ is the training set.
%---------------------------------------------------
\smallskip
\\\noindent {\bf Step 2: }We convert a common transcription-style dataset into a pseudo-rich transcription-style dataset.
The probabilities of textual and phenomenon tokens are estimated from input speech and a style token that indicates the common or rich transcription-style dataset.
The generative probabilities of mixed textual and pseudo-phenomenon tokens $\bar{\bm{W}}$ are estimated with the trained parameters mentioned in {\bf Step 1} by using following criterion:
\begin{align}
  \bar{\bm{W}} = \argmax_{\bm{W}} P(\bm{W}|\bm{X}, s={\tt [rich]}, \hat{\bm{\Theta}}).
\end{align}
This process is performed on all the data in common transcription-style dataset.
The generated pseudo-rich transcription-style dataset with the same acoustic features as common transcription-style dataset is written as
$\mathcal{D}_{\rm pr} = \{(\bar{\bm{W}}^{1}, \bm{X}^{1}_{c}), \cdots, (\bar{\bm{W}}^{K_c}, \bm{X}^{K_c}_c)\}$.
%--------------------------------------------------
%\subsubsection{Step3}
\smallskip
\\\noindent {\bf Step 3: }At this point, we can use three datasets: common transcription-style dataset $\mathcal{D}_{\rm c}$,
rich transcription-style dataset $\mathcal{D}_{\rm r}$ and pseudo-rich transcription-style dataset $\mathcal{D}_{\rm pr}$.
The model parameters $\bm{\Theta}$ are optimized by minimizing the cross entropy loss function with the three datasets as
\begin{multline}
  {\tilde{\bm{\Theta}}} = \argmin_{\bm{\Theta}} \\- \sum_{\mathcal{D} \in {\{\mathcal{D}_{\rm c},\mathcal{D}_{\rm r},\mathcal{D}_{\rm pr}\}}} \sum_{(\bm{W}^{\prime}, \bm{X}^{\prime}) \in {\mathcal{D}}} \log P(\bm{W}^{\prime}|\bm{X}^{\prime}, s; \bm{\Theta}),
\end{multline}
\begin{subnumcases}
  {s =}
  {\tt [rich]} \text{~ ~ ~ if $\mathcal{D} = \mathcal{D}_{\rm r}$ or $\mathcal{D}_{\rm pr}$,} & \\
  {\tt [common]} \text{~ ~ ~  if $\mathcal{D} = \mathcal{D}_{\rm c}$.} &
\end{subnumcases}
%In the experiments, we also investigate a condition of using only $\mathcal{D}_{r}$ and $\mathcal{D}_{pr}$ with giving ${\tt [rich]}$ and ${\tt [common]}$ respectively.
% -----------------------------------------------------------------
\subsection{Decoding}
In the decoding, beam search decoding is conducted with the trained parameters ${\tilde{\bm{\Theta}}}$ mentioned in {\bf Step 3}.
We use the following criterion during beam-search decoding with a rich-style style token:
\begin{align}
  \hat{\bm{W}} = \argmax_{\bm{W}} P(\bm{W}|\bm{X}, s= {\tt [rich]}; \bm{\tilde{\Theta}}).
\end{align}
The model can output both textual and phenomenon tokens by receiving the style token ${\tt [rich]}$.
% -----------------------------------------------------------------
\section{Experiments}
\label{sec:exp}
\subsection{Setups}
\begin{table}[t]
\centering
\caption{Details of datasets.}
\vs\vs
\label{tab:data}
\begin{tabular}{ll|rrr}
\hline
Data                   & Corpus & \begin{tabular}[c]{@{}r@{}}\# of \\ textual\\ tokens\end{tabular} & \begin{tabular}[c]{@{}r@{}}\# of \\ phenomenon\\ tokens\end{tabular} & hours \\ \hline
\multirow{2}{*}{Train} & NTDC   & 548,886                                                           & 116,906                                                              & 22    \\
                       & CSJ    & 12,568,607                                                        & -                                                                    & 545   \\ \hline
\multirow{2}{*}{Dev}   & NTDC   & 21268                                                             & 4,083                                                                & 1     \\
                       & CSJ    & 68458                                                             & -                                                                    & 2     \\ \hline
Eval                   & NTDC   & 22579                                                             & 4387                                                                 & 1     \\ \hline
\end{tabular}
\vs
\vs\vs\vs
\end{table}
%\section{Experiments}
%\label{sec:exp}
%\svs
%\subsection{Setups}
%\svs
\subsubsection{Dataset}
Table \ref{tab:data} shows details of the training, development, and evaluation sets.
We used two corpora: the corpus of spontaneous Japanese (CSJ) \cite{CSJ} as the common transcription-style dataset and
NTDC as the rich transcription-style dataset.
We split NTDC into 22 hours for the training set, 1 hour for the development set, and 1 hour for the evaluation set.
The details of NTDC is described in Section \ref{sec:data}.
For CSJ, we used 545 hours for the training set and 2 hours for the development set.

\subsubsection{ASR systems}
We use 40-dimensional log Mel-filterbank with delta and acceleration coefficients as the acoustic features.
We applied SpecAugment \cite{spaug} to the training data.
The vocabulary size was 3307 characters including textual tokens, phenomenon tokens,
and style tokens for common transcription-style and rich transcription-style.
The RT-ASR system was a transformer encoder-decoder.
The encoder and decoder each had six transformer blocks.
The token embedding dimension, hidden state dimension, non-linear layer dimension, and the number of heads were 256, 256, 2048, and 4, respectively.
The acoustic features were transformed using a two-layers 2D convolutional neural network. We set the mini-batch size to 32. For regularization, we use label smoothing \cite{label_smooth} with a smoothing parameter of 0.1 and set the dropout rate in the transformer blocks to 0.1 \cite{JMLR:v15:srivastava14a}. We used Adam optimizer \cite{adam} with Noam learning rate scheduler with 25000 warmup steps.
Early stopping was applied if no best model was found in the development set for five epochs. Gradient clipping with a maximum norm of 5.0 was applied.
When decoding by beam search, the beam size was set to 20.
We prepared a general end-to-end ASR system with a transformer encoder-decoder for comparison with our RT-ASR system.
The general end-to-to ASR system had the same configurations as the RT-ASR system.
In the training of the general ASR system, the phenomenon tokes were deleted from the NTDC.
\subsection{Results}
\begin{table*}[t]
\centering
\caption{\%CERs in different training data and input style tokens. C, R, and PR mean common transcription-style, rich transcription-style, and pseudo rich transcription-style datasets respectively. The style tokens are shown in parentheses in each dataset. SSL denote semi-supervised learning. Column of style token indicates whether or not a style token was used for models.}
\svs
\vs\vs\vstmp
\label{tab:cer}
\begin{tabular}{llllrr}
\hline
System                   & \begin{tabular}[c]{@{}l@{}}Proposed\\ SSL\end{tabular} & \begin{tabular}[c]{@{}l@{}}Training\\ Data\end{tabular} & Style-Token & \begin{tabular}[c]{@{}r@{}}\%CER\\ w/o phenomena\end{tabular} & \begin{tabular}[c]{@{}r@{}}\%CER\\ w/ phenomena\end{tabular} \\ \hline
\multirow{2}{*}{ASR}    & -                                                      & C                                                       & -           & 24.7                                                          & 31.2                                                         \\
                        & -                                                      & C+R                                                     & -           & 23.5                                                          & 34.1                                                         \\ \hline
\multirow{2}{*}{RT-ASR} & no                                                     & R                                                       & no          & 46.7                                                          & 48.2                                                         \\
                        & no                                                     & C+R                                         & yes         & 23.2                                                          & 26.8                                                         \\
                        & yes                                                    & C+R+PR                        & yes         & $\bm{22.7}$                                                          & $\bm{26.2}$                                                         \\ \hline
    \end{tabular}
    \vs
\end{table*}
\begin{figure}[t]
  \centering
  \vs\vs\vs\vs\vs\vs\vs\vs\vs\vs\vs\vs\vs\vs\vs\vs\vs
  \includegraphics[width=\linewidth]{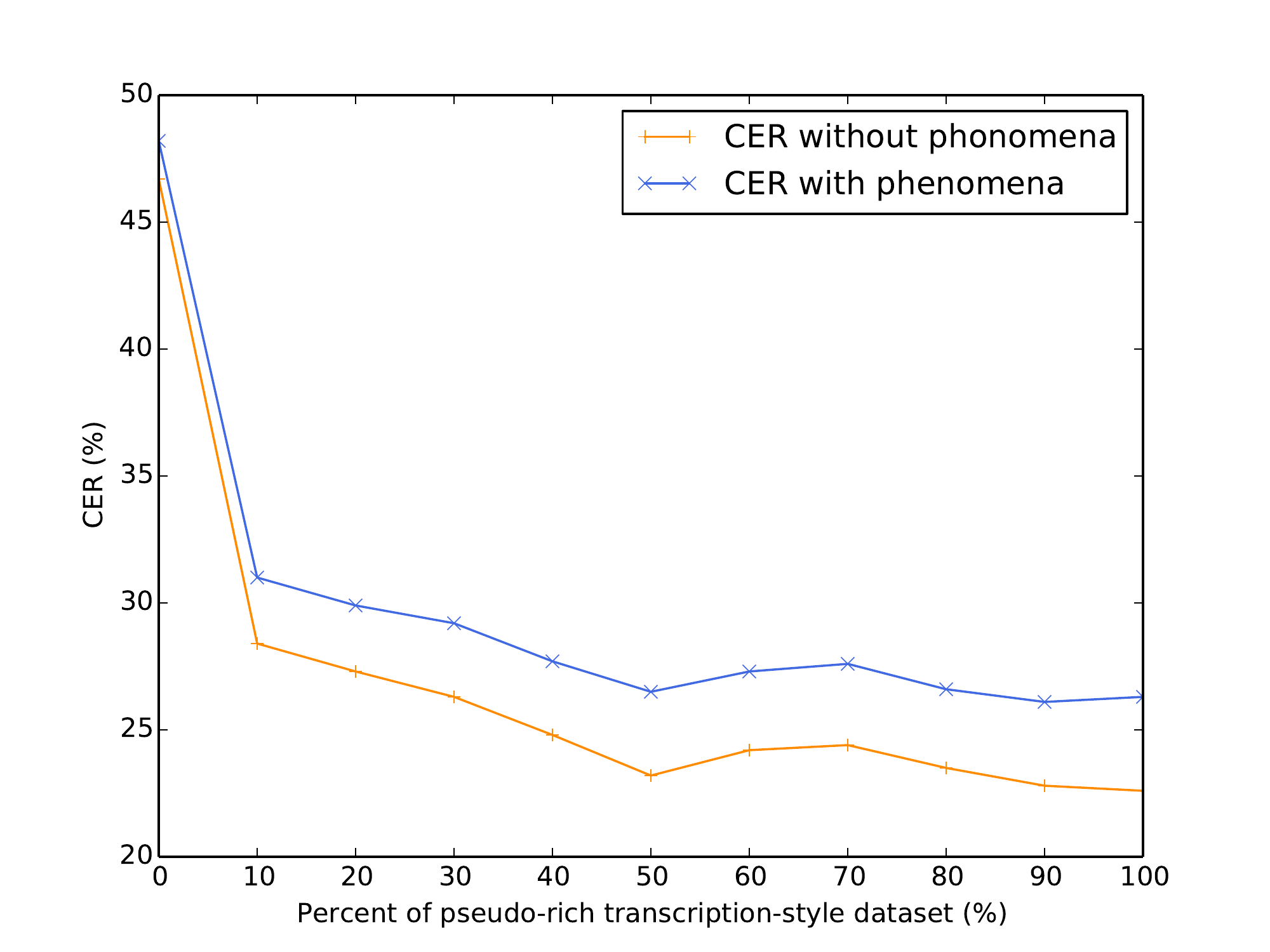}
  \vs\vs\vs\vs\vs\vs\vs\vs\vs\vs\vs\vs\vs
  \vstmp\vstmp\vstmp\vstmp\vstmp\vstmp\vstmp\vstmp
  \caption{CERs on evaluation set with different sizes of data of pseudo-rich transcription-style dataset.}
  \label{fig:cer}
  \vs\vs\vs\vs\vs\vs
\end{figure}
Table \ref{tab:cer} shows the speech recognition accuracies of the general end-to-end ASR system and the RT-ASR system in terms of CER with different training data.
Evaluation was conducted using general CER for textual tokens and the CER for textual and phenomenon tokens.
The results show that the RT-ASR system was more accurate than the general ASR system when the same common and rich transcription-style datasets (C+R) were used.
This indicates that the style tokens allowed us to explicitly consider the transcription-styles as the contexts.
The RT-ASR system with our proposed semi-supervised learning method showed better CER than the RT-ASR system trained from only rich transcription-style dataset (R).
It was difficult to build an RT-ASR system with high accuracy when only the rich transcription-style dataset was used, but was easy with the pseudo-rich transcription-style dataset.
The RT-ASR system with our proposed semi-supervised learning method outperformed the general ASR and RT-ASR systems trained from common and rich transcription-style datasets (C+R).
This system was trained from the same data as the system trained from common and rich transcription-style datasets (C+R) because pseudo-rich transcription-style dataset was generated from the common transcription-style dataset.
This confirmed the effectiveness of our semi-supervised learning method.

We constructed models by changing the size of the pseudo-rich transcription-style dataset to confirm the effect of the size of this dataset on ASR performance.
Figure \ref{fig:cer} shows the CERs with different sizes of the pseudo-rich transcription-style dataset.
The speech recognition accuracy improved as the dataset size increased from the case of 0\% on the horizontal axis (only using the rich transcription-style dataset).
These results confirmed that highly accurate RT-ASR systems can be built even with half the amount of the pseudo-rich transcription dataset.
\section{Conclusion}
\label{sec:conc}
We proposed a semi-supervised learning method for building RT-ASR systems from small-scale rich transcription-style dataset and large-scale common transcription-style datasets.
With our semi-supervised learning method, we generate a pseudo-rich transcription-style dataset from a common transcription-style dataset.
We introduced style tokens indicated rich transcription-style or common transcription-style to handle the two types of datasets efficiently.
By considering speech phenomena in spontaneous speech and applying our semi-supervised learning method, the resulting RT-ASR system outperformed a general end-to-end ASR system with the same training data.
%\bibliographystyle{IEEEtran}
%\bibliography{refs}

\end{document}